%% file: main.tex
\documentclass[conference]{IEEEtran}
\IEEEoverridecommandlockouts
\usepackage{amsmath,amssymb,amsfonts}
\usepackage{algorithm}
\usepackage{hyperref}
\usepackage{comment}
\usepackage{algpseudocode}
\usepackage{graphicx}
\graphicspath{{./images/}}
\usepackage{textcomp}
\usepackage{xcolor}
\def\BibTeX{{\rm B\kern-.05em{\sc i\kern-.025em b}\kern-.08em
    T\kern-.1667em\lower.7ex\hbox{E}\kern-.125emX}}
\usepackage
[maxbibnames=99,
firstinits=true,
sorting=none,
backend=biber]
{biblatex}
\hypersetup{
    colorlinks=true,
    linkcolor=black,
    filecolor=black,
    citecolor=black,
    urlcolor=blue,
    }
\begin{document}

\title{TRASH: Tandem Rover and Aerial Scrap Harvester\\
}

\author{\IEEEauthorblockN{Lee Milburn}
\IEEEauthorblockA{\textit{College of Engineering} \\
\textit{Northeastern University}\\
Boston, Massachusetts \\
milburn.l@northeastern.edu}
\and
\IEEEauthorblockN{John Chiaramonte}
\IEEEauthorblockA{\textit{College of Engineering} \\
\textit{Northeastern University}\\
Boston, Massachusetts \\
chiaramonte.j@northeastern.edu}
\and
\IEEEauthorblockN{Jack Fenton}
\IEEEauthorblockA{\textit{College of Engineering} \\
\textit{Northeastern University}\\
Boston, Massachusetts \\
fenton.j@northeastern.edu}
\and
\IEEEauthorblockN{Taskin Padir}
\IEEEauthorblockA{\textit{College of Engineering} \\
\textit{Northeastern University}\\
Boston, Massachusetts \\
t.padir@northeastern.edu}
}

\maketitle

\begin{abstract}
Addressing the challenge of roadside litter in the United States, which has traditionally relied on costly and ineffective manual cleanup methods\footnote{https://www.conserve-energy-future.com/littering-effects-humans-animals-environment.php}, this paper presents an autonomous multi-robot system for highway litter monitoring and collection. Our solution integrates an aerial vehicle to scan and gather data across highway stretches with a terrestrial robot equipped with a Convolutional Neural Network (CNN) for litter detection and mapping. Upon detecting litter, the ground robot navigates to each pinpointed location, re-assesses the vicinity, and employs a "greedy pickup" approach to address potential mapping inaccuracies or litter misplacements. Through simulation studies and real-world robotic trials, this work highlights the potential of our proposed system for highway cleanliness and management in the context of Robotics, Automation, and Artificial Intelligence.
\end{abstract}


\section{Introduction}
\input{sections/1_introduction}

\section{Related Work}
\input{sections/2_related_work}

\section{System Design}
\input{sections/3_system_design}

\section{Experiments}
\input{sections/4_experiment}

\section{Results}
\input{sections/5_results}

\section{Conclusion}
\input{sections/6_conclusion}

\section{Acknowledgments}

\input{sections/7_acknowledgments}

\input{sections/8_references}
\end{document}

%% file: sections/1_introduction.tex
Roadside littering poses significant environmental and aesthetic challenges. Across the US national highways, there is an estimated 51.2 billion pieces of litter\footnote{https://kab.org/litter/litter-study/}. Addressing this concern, municipalities like Monterey County Public Works and Caltrans allocate substantial resources, amounting to an annual expenditure of \$50 million\footnote{https://www.montereyherald.com/2021/01/23/roadside-trash-a-growing-problem/}. Despite these efforts, the litter problem persists, suggesting the need for innovative solutions.

\begin{figure}
\centering
\includegraphics[width=0.48\textwidth]{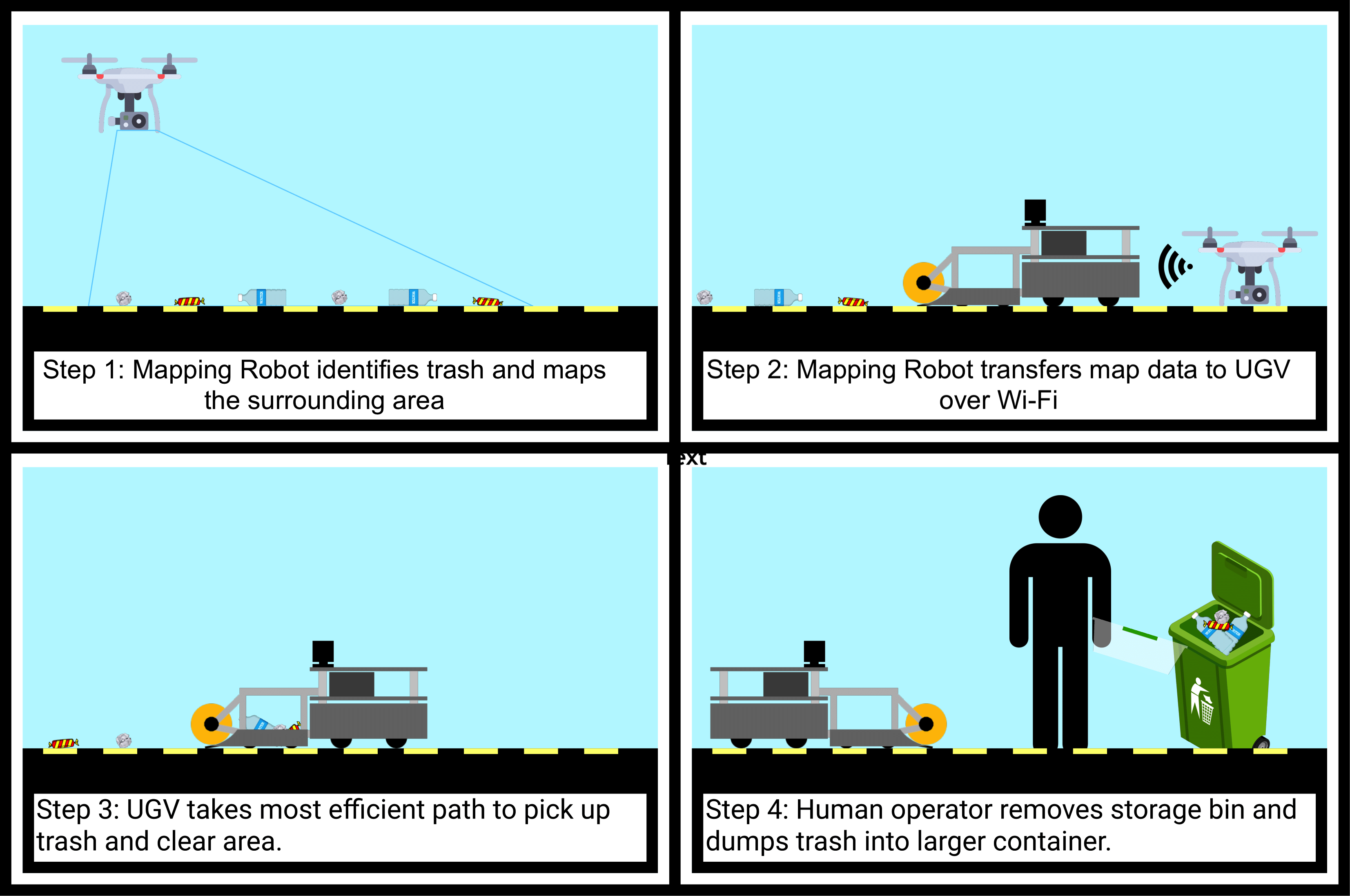}
\caption{System Design Overview.}
\label{fig:Overview}
\end{figure}

To address this challenge, we introduce an autonomous multi-robot system that employs a three-stage approach: Mapping, Navigation, and Greedy Pickup. The procedure initiates with a drone, surveying a designated road area. The acquired visuals are wirelessly transmitted to a ground robot, facilitating the creation of a comprehensive map and the pinpointing of litter locations. Subsequently, the ground robot navigates to each litter spot, employing the greedy pickup mode for scanning, re-identification, and collection. This method anticipates and rectifies potential inaccuracies in mapping and identification, ensuring an effective pickup without the necessity for extensive processing power or high-end sensors. The proposed framework is depicted in Fig. \ref{fig:Overview}.

This paper's primary contributions are:

\begin{itemize}
\item The design, development, and validation of an autonomous multi-robot \href{https://github.com/Capstone-W3/trash_parent_repo}{system}, integrating open-source mapping and vision software, bolstered by custom algorithms and hardware for trash collection.
\item Introduction of algorithms that actively rectify inaccuracies in the system's procedures during runtime, notably the Greedy Pickup Procedure and the anti-clustering algorithm.
\item The formulation of a manually annotated \href{https://app.roboflow.com/trash-northeastern/trash-neu-original/}{dataset} encompassing over 1,250 images, curated for optimized trash object detection.
\end{itemize}

%% file: sections/2_related_work.tex
Multi-robot systems have been extensively studied and utilized for various environmental monitoring tasks. Espina et al. \cite{Espina2011} explored the efficacy of multi-robot systems in environmental monitoring, concluding that these systems present a more effective solution than static monitoring methods. There has also been a surge in interest in using autonomous robots for trash collection \cite{9561809}. The key finding here was that robots should be environment-aware rather than relying solely on their field of view (FOV) for trash collection, ensuring a more efficient trash collection process.

There are other notable applications of multi-robot systems, such as search and rescue \cite{multiRobotSearchRescue}, forest fire prevention \cite{multiRobotForestFires}, and construction \cite{multiRobotConstruction}. While these frameworks are established and effective, their application to roadside trash collection remains a largely unexplored domain.

Several trash collection robotic systems have been designed for specific terrains such as rivers or the ground. Some of these include \cite{GrassBot},\cite{BeachBot}, and \cite{riverBot}. Most of these robots are either part of homogeneous multi-robot systems, are not autonomous, or operate individually. Furthermore, they do not autonomously identify and store trash as waypoints. Instead, they opt for a more myopic approach, searching for and collecting the closest piece of trash. This method doesn't guarantee thoroughness since there is no initial estimate of the total trash amount in an area. In contrast, our system, as detailed in this paper, offers a comprehensive approach to tackle the roadside trash problem.

An autonomous system designed to collect and monitor plastics in rivers was introduced by \cite{jmse9080879}. This system integrates a central processor that assigns tasks to underwater autonomous vehicles that then pick up plastics. The paper concluded that a Multi-robot task allocation architecture \cite{inbook} with a central control mechanism enhances the system's efficiency. However, there is room for improvement in the hardware used in such environments.

A summarized comparison of several existing multi-robot and trash collection systems can be seen in Table \ref{table:comparison}. As can be observed, the system proposed in this paper offers advantages in terms of autonomy, path planning, adaptability to various terrains, use of machine learning, and cost-effectiveness, making it a comprehensive solution to the roadside trash problem.

\begin{table}[h]
    \centering
    \caption{Comparison of various trash collection systems.}
    \label{table:comparison}
    \small 
    \setlength{\tabcolsep}{1.5pt} 
    \begin{tabular}{|l|c|c|c|c|c|c|c|c|c|}
        \hline
        \textbf{Ref.} & \textbf{Auto.} & \textbf{Multi-R.} & \textbf{Hetero.} & \textbf{Path} & \textbf{Multi-T} & \textbf{ML} & \textbf{Cost-Eff.} & \textbf{Trash}\\
        \hline
        \cite{multiRobotSearchRescue} & Y & Y & Y & Y & Y & Y & N & N \\
        \cite{multiRobotForestFires} & N & Y & N & Y & Y & N & N & N \\
        \cite{multiRobotConstruction} & Y & Y & Y & Y & Y & N & N & N \\
        \cite{GrassBot} & Y & N & N & Y & N & Y & Y & Y\\
        \cite{BeachBot} & N & Y & N & N & N & N & Y & Y\\
        \cite{riverBot} & Y & N & N & Y & N & Y & Y & Y\\
        \cite{beach_radio} & N & N & N & N & Y & N & Y & Y\\
        \cite{brooms} & Y & N & N & Y & N & Y & N & Y\\
        \cite{hsr_cleaning} & Y & N & N & Y & N & Y & N & Y\\
        \cite{bansal2019agdc} & Y & N & N & Y & N & Y & N & Y\\
        \cite{IWSCR} & Y & N & N & Y & N & Y & Y & Y\\
        \cite{electronics10182292} & Y & N & N & Y & Y & Y & Y & Y\\
        \textbf{This} & \textbf{Y} & \textbf{Y} & \textbf{Y} & \textbf{Y} & \textbf{Y} & \textbf{Y} & \textbf{Y} & \textbf{Y}\\
        \hline
    \end{tabular}
    \vspace{1em} 
    \begin{flushleft} 
        \textbf{Legend:}
        \small
        \begin{tabular}{ll}
            \textbf{Auto.} & Autonomous \\
            \textbf{Multi-R.} & Multi-Robot \\
            \textbf{Hetero.} & Heterogeneous \\
            \textbf{Path} & Path Planning \\
        \end{tabular}
        \quad 
        \begin{tabular}{ll}
            \textbf{Multi-T} & Multi-Terrain \\
            \textbf{ML} & Machine Learning \\
            \textbf{Cost-Eff.} & Cost Effective \\
             \textbf{Trash} & Collects Trash \\
        \end{tabular}
    \end{flushleft}
\end{table}

There have also been works into advancing ground robots traversing and collecting items in hard terrain, \cite{watanabe2020study}, \cite{buczylowski2020handheld}. These works show effective ways that the ground robot can move and collect trash in challenging roadside environments. 

For our system to be cheap and lightweight, we integrated existing software was needed that works in real-time on standard CPUs in a wide variety of environments. The Robotic Operating System (ROS) \cite{Quigley09} is an open-source robotics framework that allows each of our hardware and software components to communicate freely in real-time, and each software component used was compatible with this framework. ORB-SLAM2 was the ideal solution for mapping \cite{github-raulmur}\cite{7219438}. It uses differing angles of static environmental features to create a map and a keyframe-based SLAM approach that reduces the overall data size of the SLAM map considerably \cite{AI2020}. Since the system is designed with visual sensors, software to visually identify trash was necessary. YoLOv4 is a CNN model trained from annotated images to place bounding boxes around specified objects in RBG images. Adaptive Monte Carlo Localization (AMCL) is the method of navigation used as well as the name of a compatible software stack used for navigation provided by ROS \cite{amcl}. AMCL takes in odometry feedback from the robot's wheels and scans data derived from the RGB-D Camera to navigate. After some static conversions from ORB-SLAM2's native map format to a 2D occupancy grid, AMCL can autonomously navigate around an environment. These existing software stacks served as the framework for the multi-robot system to be built.

%% file: sections/3_system_design.tex
\begin{figure}
    \centering
    \includegraphics[width=0.45\textwidth]{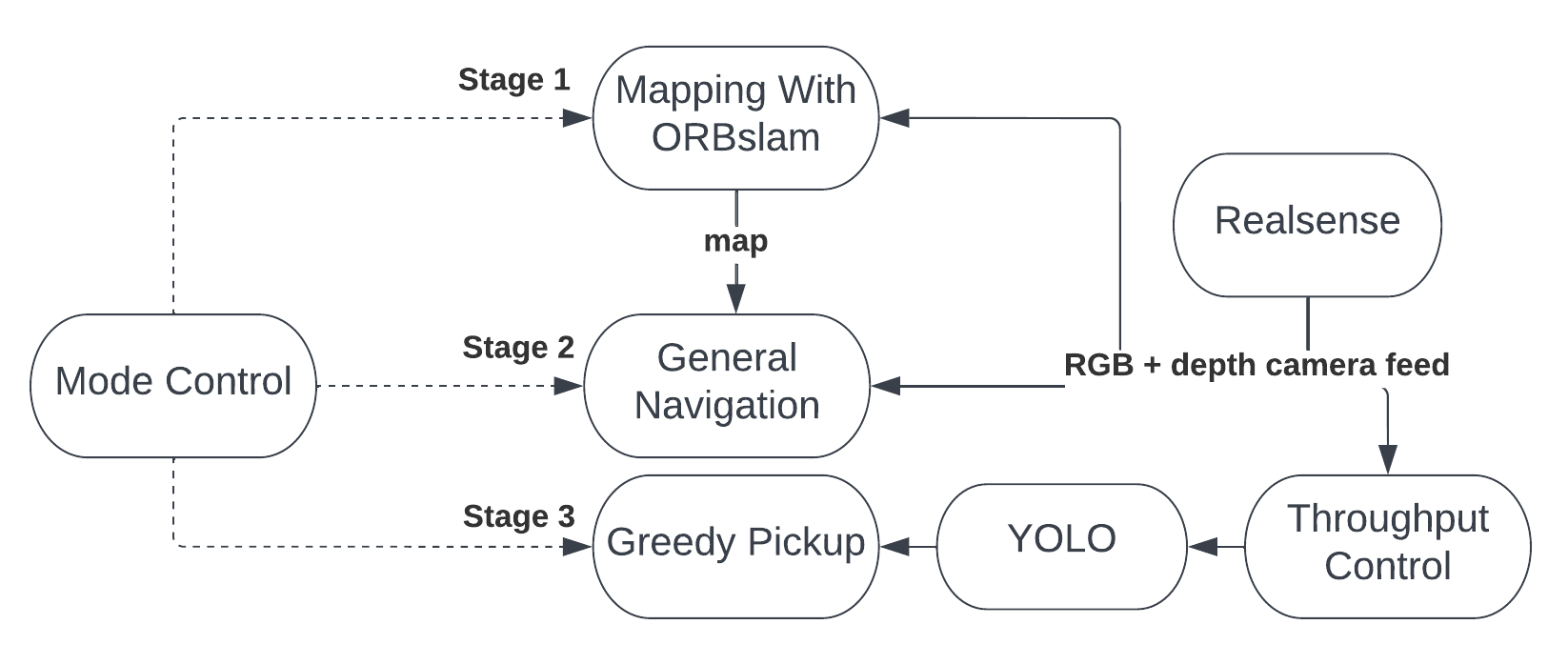}
    \caption{Systems' Communication Flow.}
    \label{fig:ComFlow}
\end{figure}

The proposed framework integrates custom algorithms with the convolutional neural network You Only Look Once (YOLOv4) for image identification \cite{7780460}, the visual SLAM solution ORBSLAM-2 for map-building \cite{7219438}, and the ROS Adaptive Monte Carlo Localization (AMCL) package \cite{amcl}, for path planning between generated trash points in the map and obstacle avoidance. 

 After the operator initializes the mapping robot, which scans an area using an RGB-D sensor. The sensor data, processed by ORB-SLAM2, produces a continuous digital map of the target area. The UGV uses YOLOv4 to detect trash and mark the trash piece's coordinates on the two-dimensional map. The UGV then plans an efficient path between the trash coordinates. As it navigates the calculated path, the UGV confirms trash locations with its RGB-D sensor and collects the identified trash. The system's integration flow is shown in Fig. \ref{fig:ComFlow}.

\begin{figure}[htbp]
  \centering
  \begin{minipage}{.24\textwidth}
    \centering
    \includegraphics[height=2.5cm]{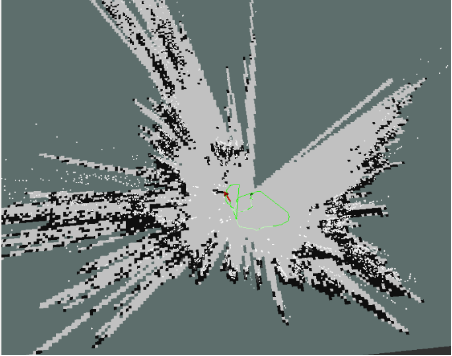}
    \caption{Raw Occupancy Grid.}
    \label{fig:OccGrid}
  \end{minipage}%
  \hfill
  \begin{minipage}{.24\textwidth}
    \centering
    \includegraphics[height=2.5cm]{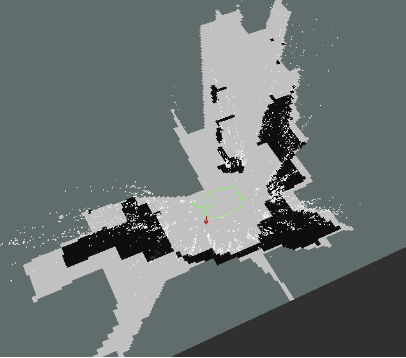}
    \caption{Morphologically Transformed.}
    \label{fig:Morph}
  \end{minipage}%
\end{figure}

\textbf{Mapping:} ORB-SLAM2 was chosen for mapping due to its keyframe-based SLAM approach, which significantly reduces the data size of the SLAM map, see Fig. \ref{fig:OccGrid}. Morphological operations clean up the image and generate smooth, continuous maps, see Fig. \ref{fig:Morph}.

\begin{figure}[h]
    \centering
    \begin{minipage}{0.24\textwidth}
        \centering
        \includegraphics[width=0.7\textwidth]{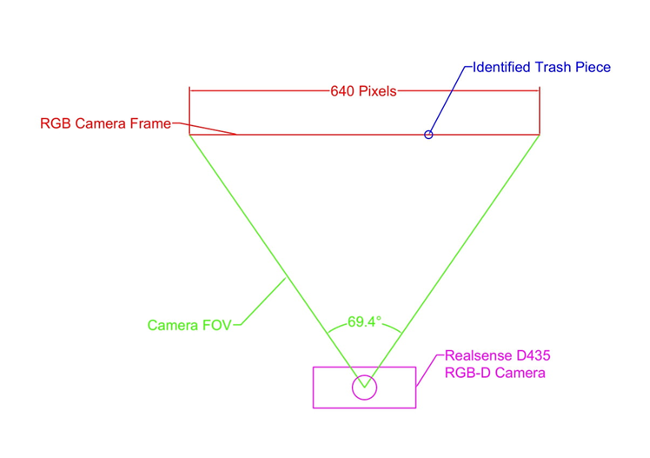} 
        \caption{FOV Diagram}
        \label{fig:FOVDia}
    \end{minipage}\hfill
    \begin{minipage}{0.24\textwidth}
        \centering
        \includegraphics[width=0.7\textwidth]{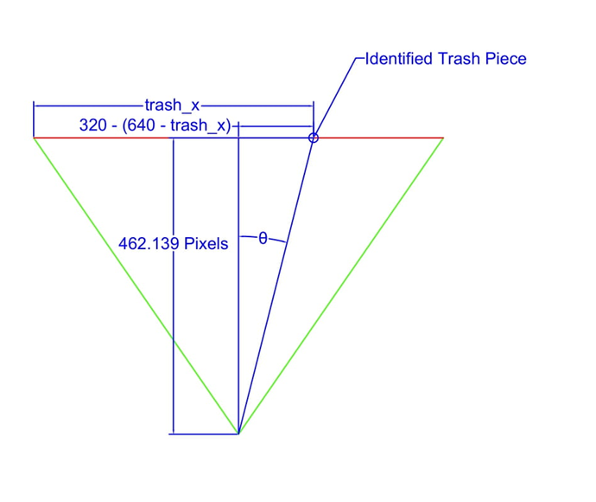} 
        \caption{FOV Trigonometric Calculations}
        \label{fig:FOVCalc}
    \end{minipage}
        \begin{minipage}{0.24\textwidth}
        \centering
        \includegraphics[width=0.5\textwidth]{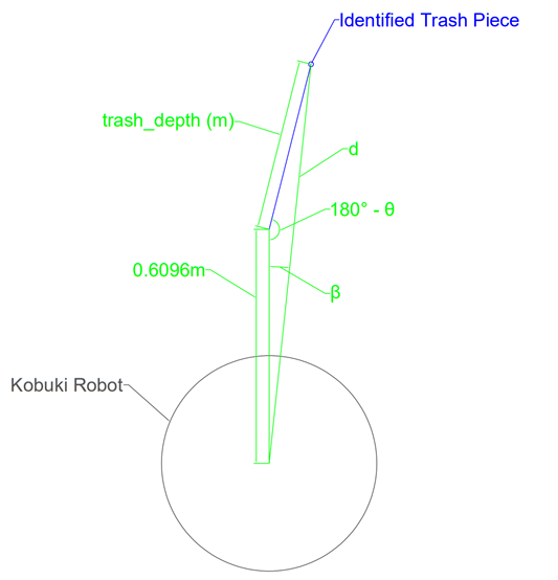} 
        \caption{Robot-Trash Trigonometric Calculations}
        \label{fig:RTCalc}
    \end{minipage}
    \begin{minipage}{0.24\textwidth}

        \centering
        \includegraphics[width=0.7\textwidth]{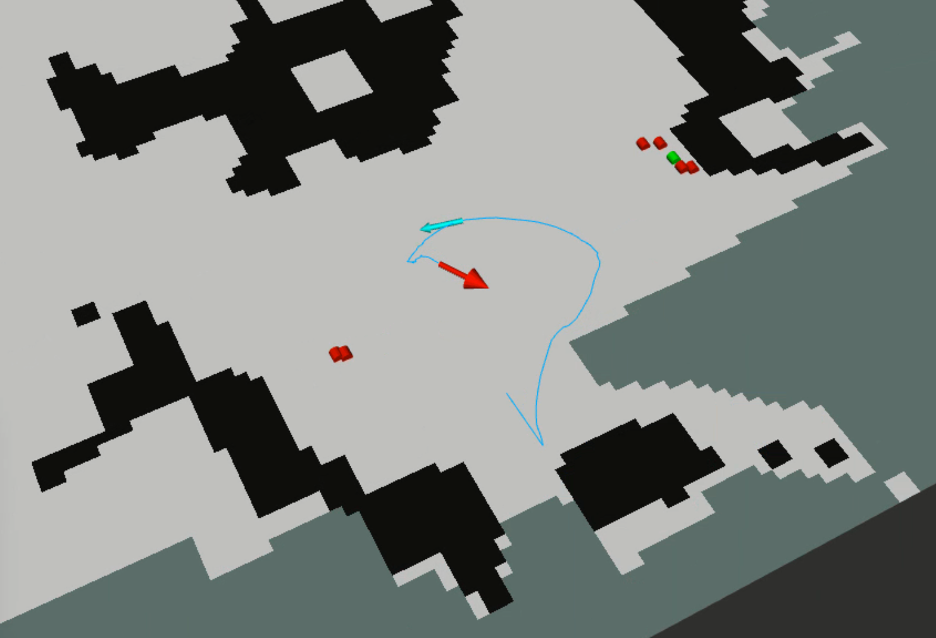}
        \caption{Robot Trajectory in the map}
        \label{fig:robTraj}
    \end{minipage}
\end{figure}

\textbf{Trash Identification:} As an area is being mapped YoLOv4 is used, a Convolutional Neural Network (CNN) model trained to identify trash trained with a custom dataset of over 1200 images of trash. The model runs simultaneously with the mapping process and provides bounding boxes around identified trash. The UavWaste dataset is used for identifying trash from an aerial camera \cite{uavWaste}. However, for the system to be cost-effective, it is created to work with low-power CPUs. As such trash is identified on a delay compared to the location of the system. Using the tf2 transformation library the time the trash detected image is taken and assigned to the location of the robot at that time in the map, known by saving the robots trajectory in the map, seen in Fig. \ref{fig:robTraj}. Then using that location, the UGV's FOV is used to determine the location of a piece of trash in relation to the robot base at that given time, as seen in Figs. \ref{fig:FOVDia}, \ref{fig:FOVCalc}, \ref{fig:RTCalc}.

\textbf{Anti-clustering:} The anti-clustering algorithm takes a rolling average of the incoming trash detection to filter redundant and noisy detections. Every time a piece of trash was detected a ROS subscriber would listen to the detection filter based on the calculations seen in the equation (\ref{eqn:antiClus}). Where $p_{1x/y}$ is the existing detection's respective x and y coordinate, $p_{2x/2y}$ is the new trash detection's x, and y coordinates and $a$ is the number of times $p_{1}$ has been averaged. 

\begin{equation}
    \label{eqn:antiClus}
    p_{1x} = \frac{p_{1x}a + p_{2x}}{a + 1} \\ 
    p_{1y} = \frac{p_{1y}a + p_{2y}}{a + 1} \\
\end{equation}

\begin{algorithm}
\caption{Anti-Clustering Algorithm}
 \hspace*{\algorithmicindent} \textbf{Input: Trash Pose $tp$} \\
 \hspace*{\algorithmicindent} \textbf{Output: Poses of clustered trash pieces $ctp$}
 \begin{algorithmic}
\State averaged Trash poses $atp$ tupled with times averaged $ta$ $(atp, ta) \gets empty$
 \If{$atp$ = 0}
 \State $atp \gets tp$
 \State $ta \gets 1$
 \Else
 \For{every pose tuple $pt$ in $atpt$}
 \State get x and y bottom and top around the trash's location 
 \State $xb, xt, yb, yt$
 \If{$xb \leq tpx \leq xt \And yx \leq tpy \leq yt$}
 \State pose tuple x $ptx$ = ($ptx$ * $ta$ + $tpx$) / ($ta$ + 1.0)
 \State pose tuple y $pty$ = ($pty$ * $ta$ + $tpy$) / ($ta$ + 1.0)
 \State $ta$ + 1
 \EndIf
 \EndFor
 \If{$tp$ is not in $atp$}
 \State $atp \gets tp$
 \EndIf
 \State $ctp \gets $ all trash poses $tp$ in $atp$ where $tp_i$'s  $ta_i > 2$ times
 \EndIf
 \end{algorithmic}
\label{alg:antiClus}
\end{algorithm}

During the initial search, the CNN model classifies random background objects as pieces of trash and gets multiple
detections of the same piece of trash. Each new trash pose detection is filtered to sort through the noisy detections. Every time
a trash piece was detected, a ROS subscriber would listen
to the detection and determine if the identified trash piece was
newly detected or detection of a trash pose had already been
found. These detections are processed in the Algorithm \ref{alg:antiClus}.

\textbf{General Navigation and Path Planning:} The robot receives the 2D occupancy grid created from ORB-SLAM2 and the trash detection coordinates. It then navigates to within two meters of the nearest detected trash point. For navigation, we used AMCL, which takes odometry feedback from the robot's wheels and scan data from the RGB-D Camera. A custom module feeds the trash detection coordinates into AMCL for path planning between trash points.

\textbf{Greedy Pickup:} Greedy Pickup is activated when the robot is within 2 meters of trash. It rotates to look for trash using YoLOv4 and calculates the position of the trash. The robot turns toward the trash, activates the collection mechanism, and moves 0.2m past the trash's location for proper collection. The handmade dataset is used for this step, as it is used for looking at trash from a ground angle as opposed to a birds-eye view, like the UAVWaste dataset. Once collected, the motor turns off, and navigation resumes. This procedure accounts for trash moving between the time it was initially identified and when it is being collected, and for false trash identifications. This approach is shown in Algorithm \ref{alg:greedy}.

\begin{algorithm}
\caption{Greedy Pickup}
 \hspace*{\algorithmicindent} \textbf{Input: Detected Trash pose list $tpl$}
 \begin{algorithmic}
 \State set timeout time $to$
 \State set confidence threshold $ct$
 \For{Trash pose $tp$ in $tpl$}
 \State Determine whether $tp$ is to the left or right of the robot in the Map
 \State Spin in direction of $tp$, scanning for confirmation
 \If{Robot gets $tp \geq ct$}
 \State Robot Stop spinning
 \State Use \textbf{tf2 transform} from trash identification to find relevant Robot orientation and trash pose  
 \State Get destination angle
 \State Robot turns to destination angle
 \State Robot Turn on collection mechanism and drive over detected piece of trash
 \Else
 \State Robot exceeded $to$ looking for trash
 \EndIf
 \EndFor
 \hspace*{\algorithmicindent} 
 
 \textbf{Output: Poses of clustered trash pieces $ctp$}
 \end{algorithmic}
\label{alg:greedy}
\end{algorithm}


%% file: sections/4_experiment.tex
\begin{figure}[htbp]
  \centering
  \hfill
  \begin{minipage}{.20\textwidth}
    \centering
    \includegraphics[height=2cm]{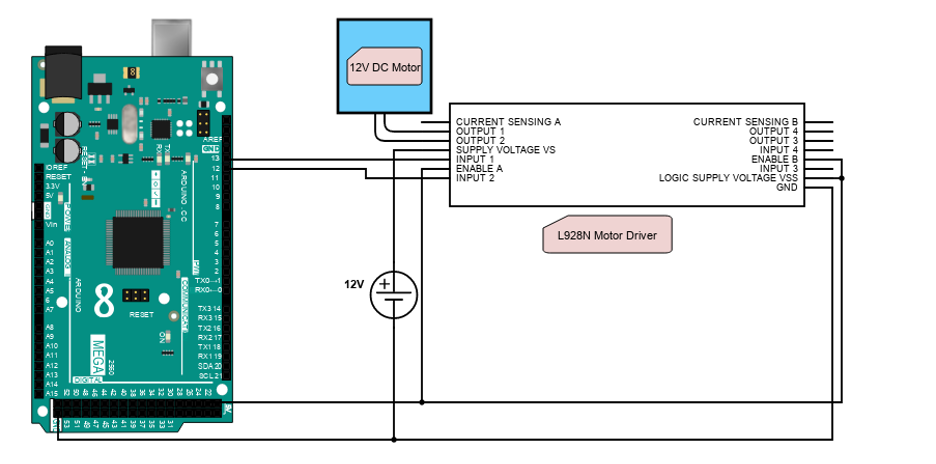}
    \caption{Motor Circuitry.}
    \label{fig:motorCirc}
  \end{minipage}
  \hfill
  \vspace{5px}
  \begin{minipage}{.2\textwidth}
    \centering
    \includegraphics[height=2cm]{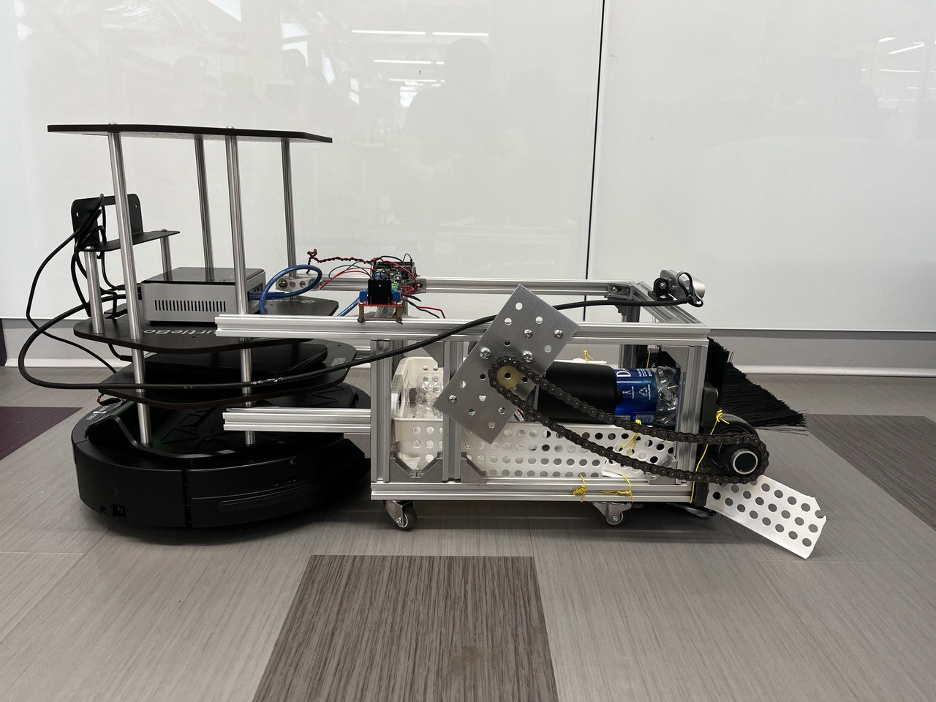}
    \caption{Full Construction.}
    \label{fig:HardwareTest}
  \end{minipage}%
  \hfill
  \begin{minipage}{.24\textwidth}
    \centering
    \includegraphics[height=2cm]{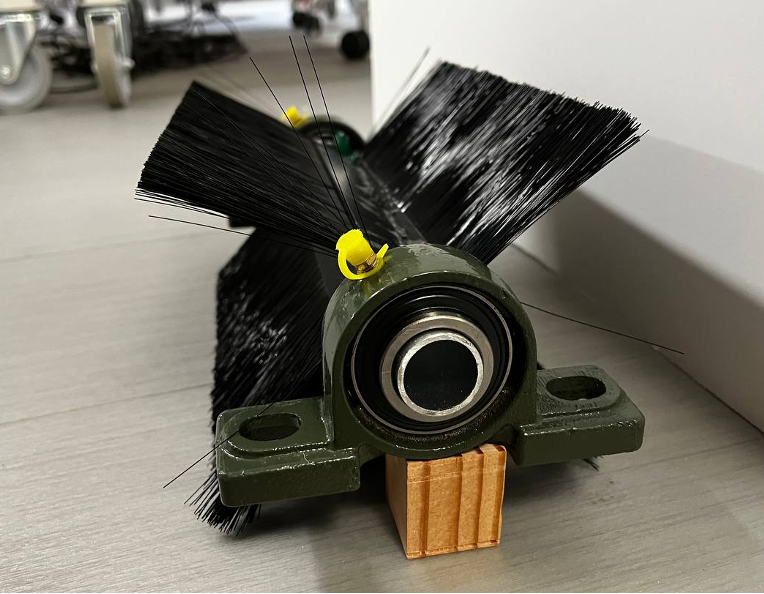}
    \caption{Custom Brush.}
    \label{fig:brush}
  \end{minipage}%
  \hfill
\begin{minipage}{.24\textwidth}
    \centering
    \includegraphics[height=2cm]{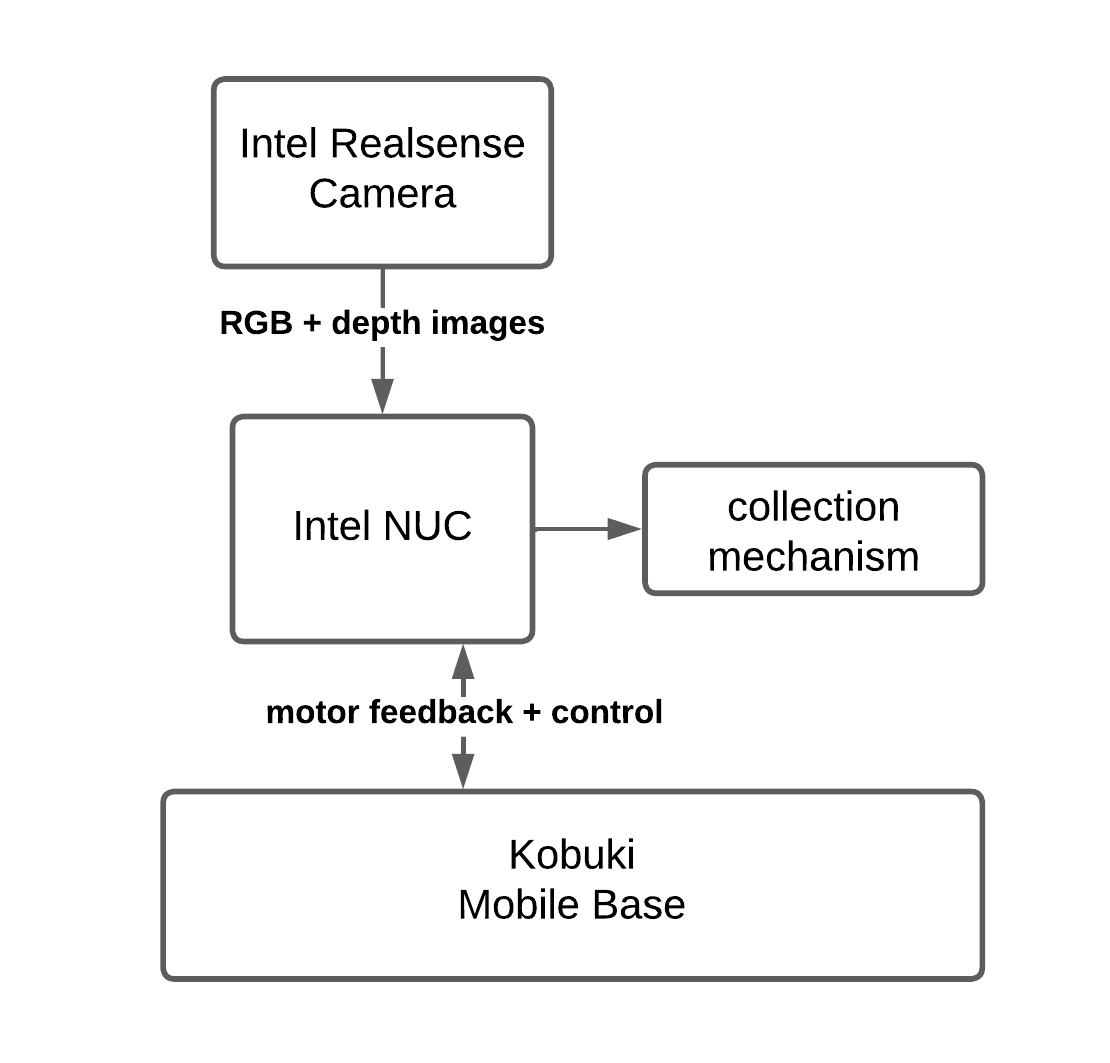}
    \caption{Hardware Overview.}
    \label{fig:HardwareOverview}
  \end{minipage}%
\vspace{-10pt} 
\end{figure}

\begin{figure}
    \centering
    \includegraphics[width=0.44\textwidth]{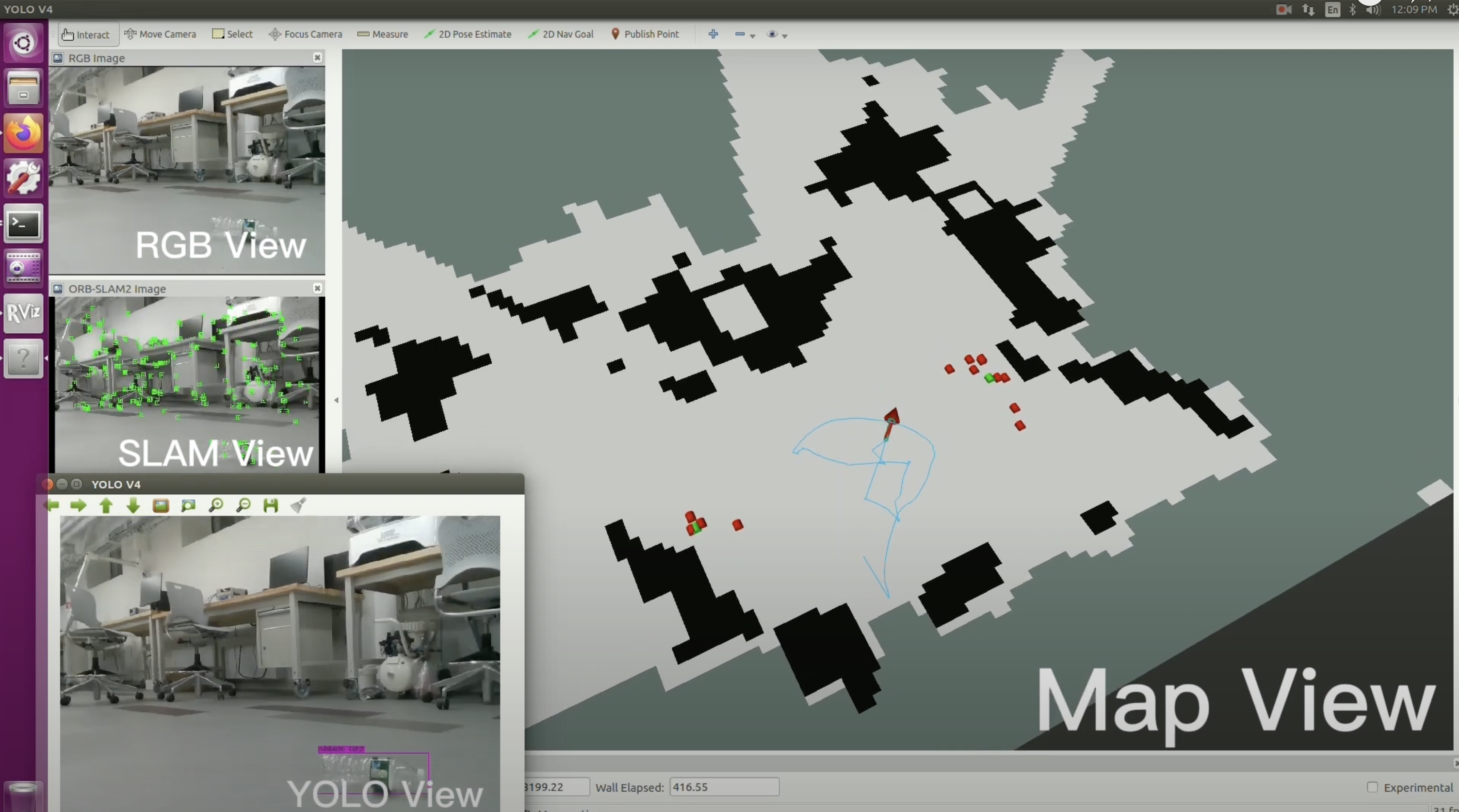}
    \caption{Real Time Trash Identification and Map Creation.}
    \label{fig:MapTest}
\end{figure}

\textbf{Goal:} These experiments were aimed at evaluating the feasibility of the system design in terms of mapping an enclosure, identifying and marking trash, choosing an efficient path, and picking up the trash. The robotic system was designed to clear trash as large and heavy as an average 600mL Spring Valley Water bottle, weighing about 0.64 kg. See Fig. \ref{fig:MapTest}.

\textbf{Testing In Simulation:} Simulated testing was conducted in the Gazebo Robotics simulator, which is included in the base Turtlebot SDK. The simulator enabled initial testing of movement and navigation.

\textbf{Hardware Setup:} The hardware design uses a modified Turtlebot-2 as a base design with four main components: Intel’s Realsense D435 RGB-D camera, an Intel NUC running a GNU-Linux OS along with ROS, a Kobuki Mobile Base, and a custom-designed collection mechanism, see Fig. \ref{fig:HardwareOverview}.

The collection mechanism's motor is controlled by sending serial packets to the Arduino. A PWM signal is output to the L928N motor controller, gradually increasing the duty cycle to avoid current spikes and ensure the motor operates within the 1.5A current limit. See the motor circuit used in Fig. \ref{fig:motorCirc}.

\textbf{The Collection Mechanism} The collection mechanism is a custom-designed addition to the Turtlebot Robot, consisting of a mechanical construction made of aluminum bars connected by 90-degree brackets. The brush was hand-designed and fabricated to fit the design specifications of the mobile robot. The electronics system for the collection mechanism is an Arduino Mega connected to the NUC using a USB cable and an L982N Motor driver breakout board. See the full construction in Fig. \ref{fig:HardwareTest}, and the custom brush in Fig. \ref{fig:brush}.

\textbf{The Environment:} The robot was tested in a room with a random configuration of chairs and obstacles, where the map would be created each time in a dynamic environment, and the system would have to account for a new configuration. Trash was placed in different locations for all tests, with up to four pieces of trash in the environment.

\textbf{Tests: }

\textbf{Testing Mapping/ Image Identification accuracy/ Accounting for Latency -} The accuracy of trash detection and associated processing latency were evaluated by comparing generated maps, the detected trash locations, against anticipated maps. A successful map is defined to accurately represent the trash positions within half a meter of their real-world location, while also including all the environmental obstacles placed.

\textbf{Testing the Greedy Pickup Algorithm -} The effectiveness of the Greedy Pickup algorithm was assessed through several trials, involving trash positioned at varying distances from the robot—2 meters, 1 meter, and half a meter. A successful trial is one in which the robot accurately identifies and collects the trash at the prescribed distance.

\textbf{Full System Trials -} The comprehensive system trials incorporated all aspects of the operation—mapping the environment, labeling trash points, navigating to labeled trash pieces, and collecting trash using the Greedy Pickup algorithm. A trial is deemed successful when the system accurately maps the environment, correctly identifies and labels trash points, efficiently navigates to each piece of trash, and successfully collects it. Future experiments aim to more thoroughly test the drone streaming information in open environments.


%% file: sections/5_results.tex
\begin{figure}[h]
\centering
\begin{minipage}{0.24\textwidth}
\centering
\includegraphics[width=0.7\textwidth]{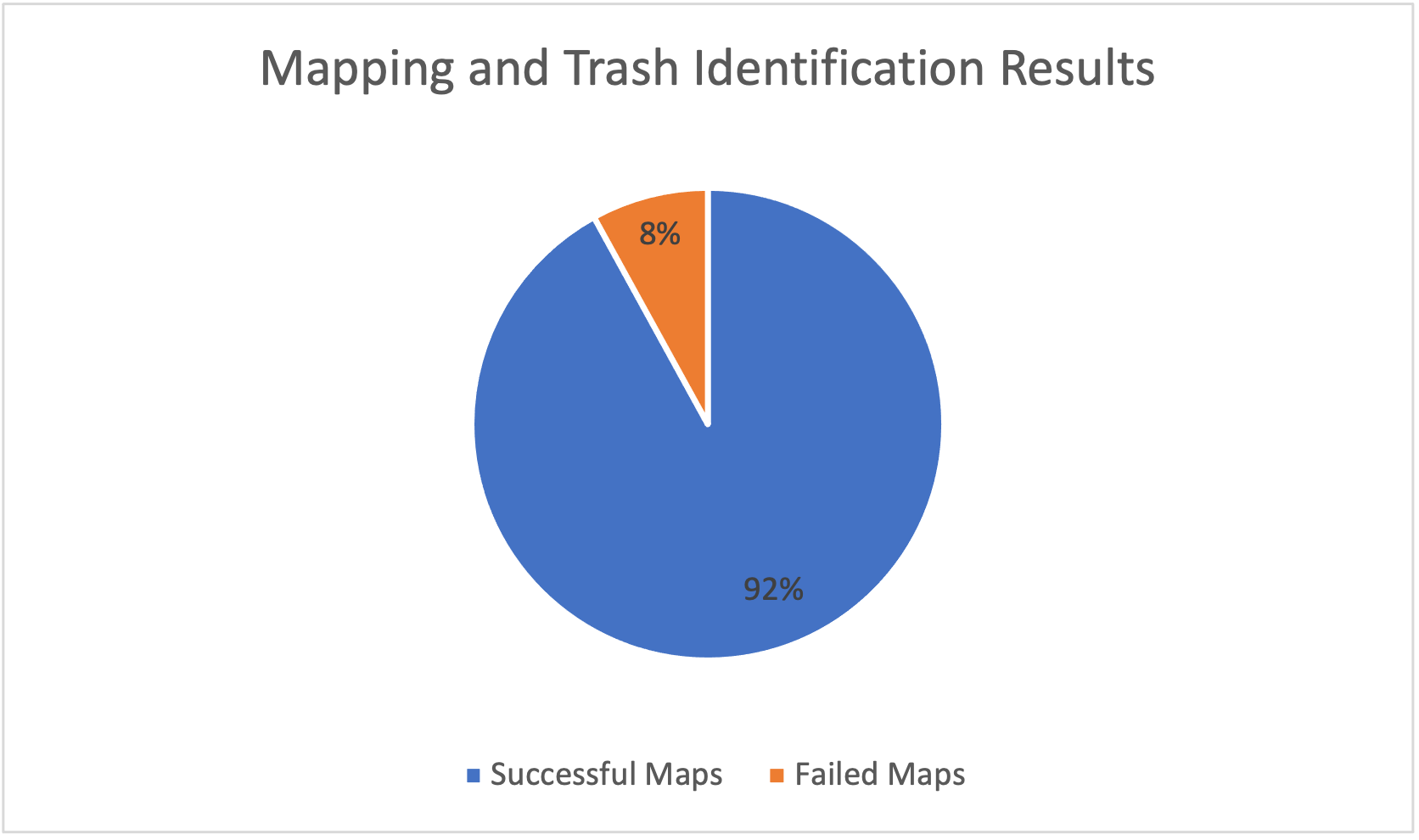}
\caption{Accurate Map Creation with Identified Trash Points Result}
\label{fig:MapResult}
\end{minipage}\hfill
\begin{minipage}{0.24\textwidth}
\centering
\includegraphics[width=0.7\textwidth]{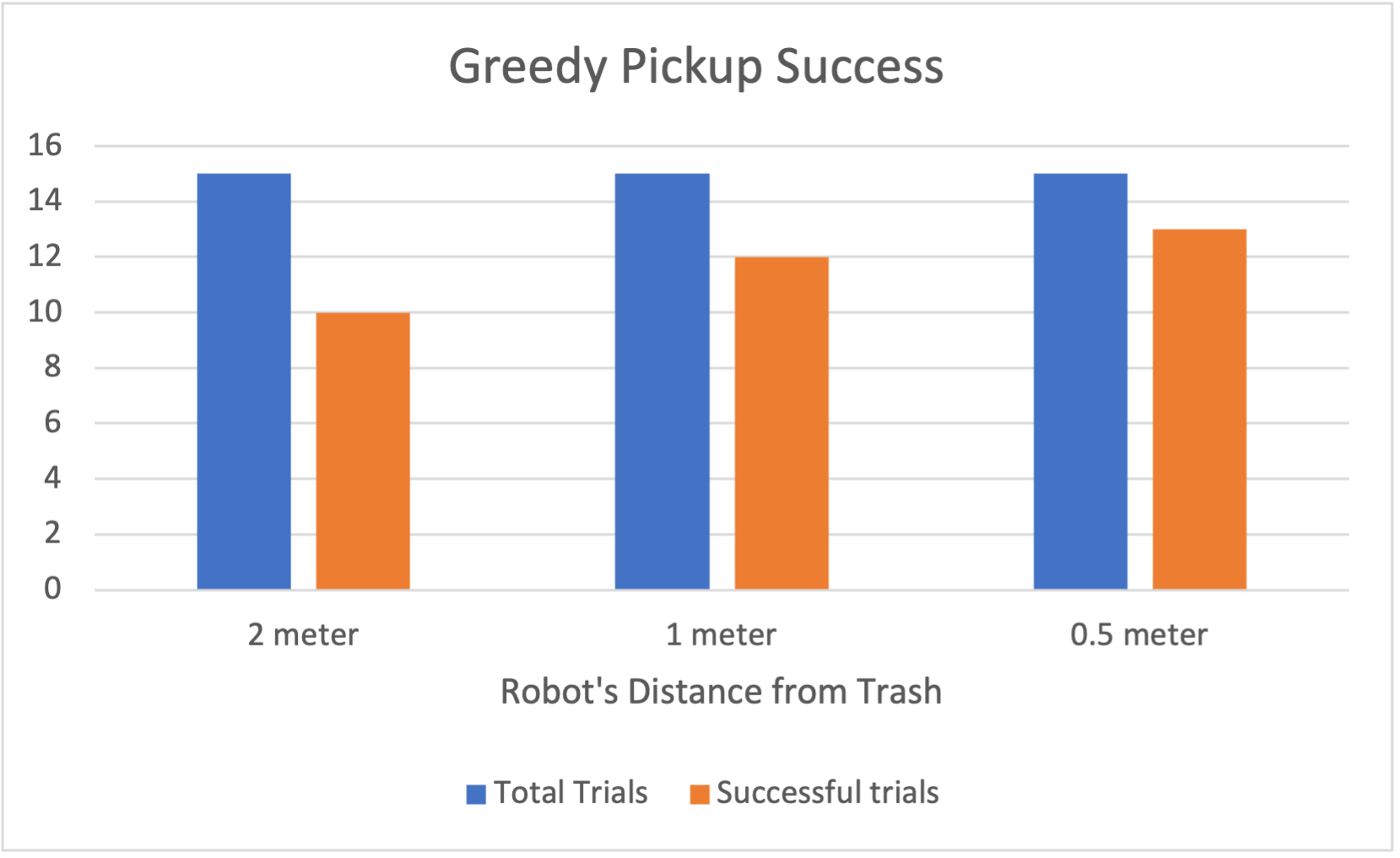}
\caption{Greedy Pickup Results}
\label{fig:GreedyPickupResult}
\end{minipage}
\begin{minipage}{0.45\textwidth}
\centering
\vspace{5pt}
\includegraphics[width=0.7\textwidth]{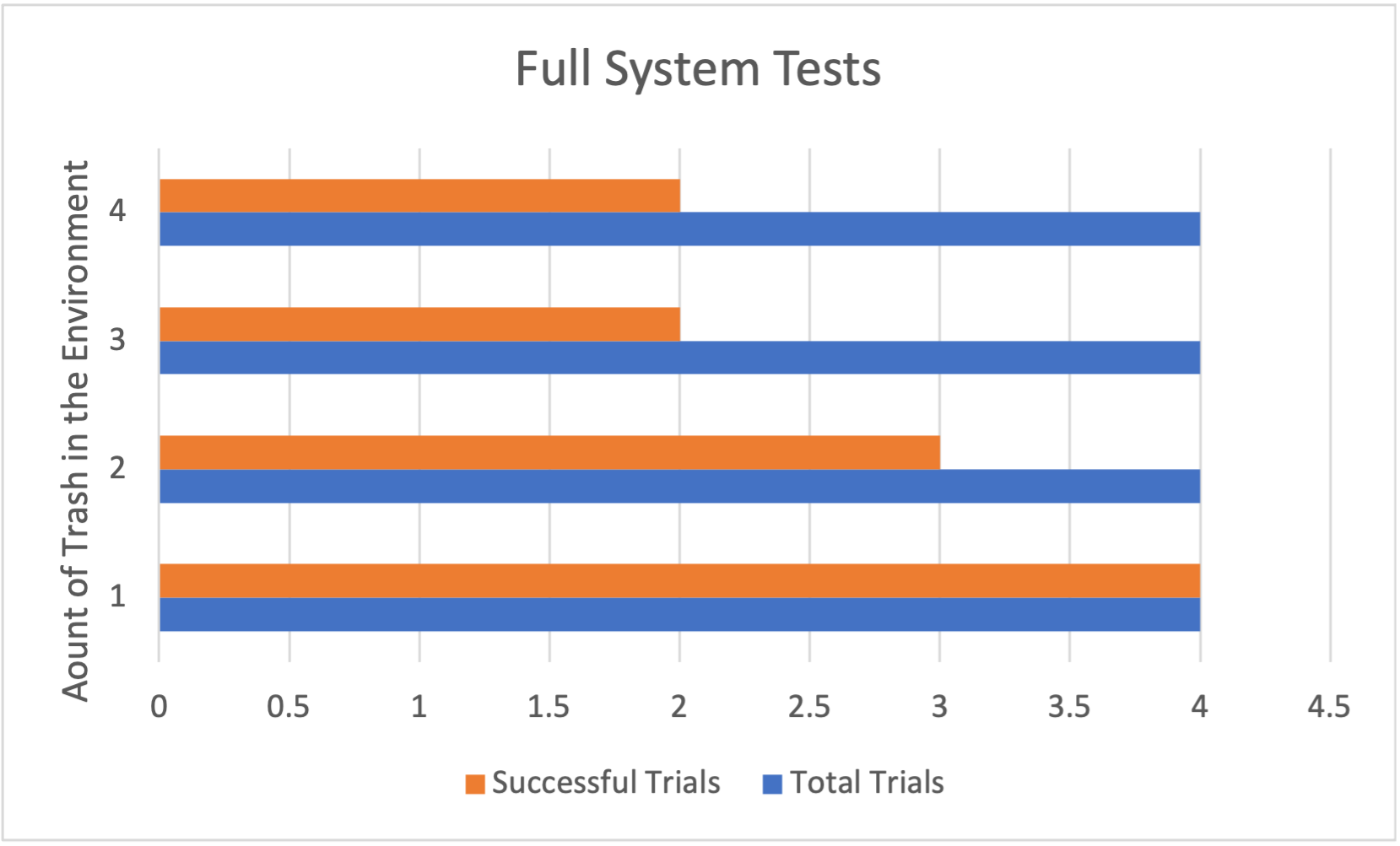}
\caption{Full System Run-through Results}
\label{fig:FullResult}
\end{minipage}
\end{figure}

\subsection{Accounting for Processing Latency}
\begin{figure}[h]
\centering
\includegraphics[width=0.44\textwidth]{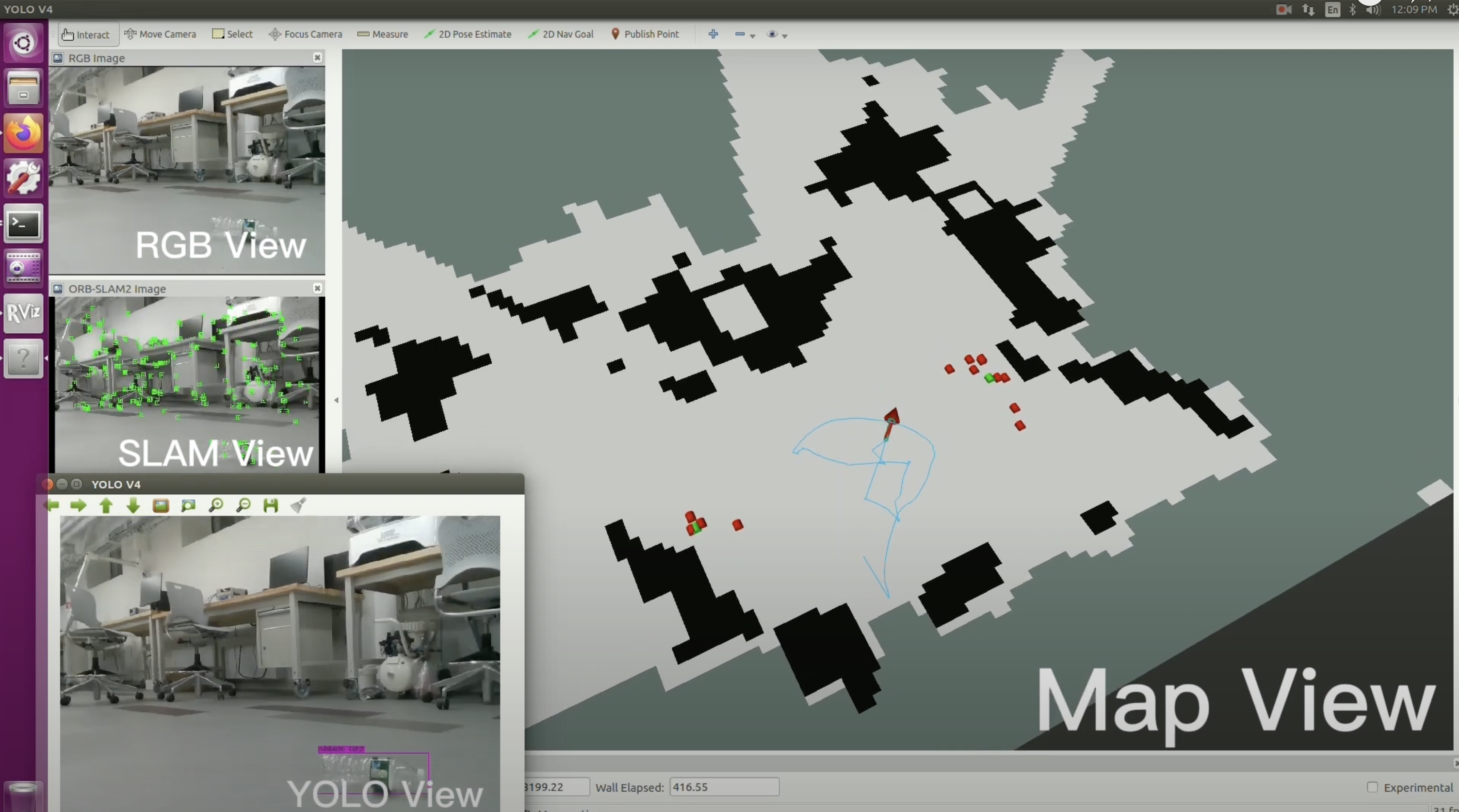}
\caption{Real Time Trash Identification and Map Creation}
\label{fig:mapResult}
\end{figure}

The map creation and image identification processes, depicted in Fig. \ref{fig:mapResult}, were effective tools for area monitoring and trash location identification. Out of 50 mapping runs conducted, the process showed significant accuracy as indicated in Fig. \ref{fig:MapResult}. Failures typically arose when ORBSLAM lost tracking and was unable to reacquire it.

\subsection{Evaluation of Greedy Pickup Algorithm}
\begin{figure}[h]
\centering
\includegraphics[height=4cm]{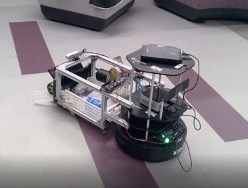}
\caption{Still of UGV Picking up Water Bottle in System Test}
\label{fig:GreedyPickup}
\end{figure}

The Greedy Pickup strategy yielded an accuracy rate of 77.78\%, as shown in Fig. \ref{fig:GreedyPickupResult}. A total of 45 trials were conducted, with 15 trials at each distance. Failures became more frequent as the distance from the trash increased, primarily due to the introduction of greater drift. See Fig. \ref{fig:GreedyPickup} for test in action.  

\subsection{Comprehensive System Performance}
The full system achieved a success rate of 68\%, as depicted in Fig. \ref{fig:FullResult}. The testing encompassed 16 runs, with four runs for each amount of trash placed in the environment. Failures predominantly occurred due to drift while attempting to pick up trash, with more possible points of failure with more trash pieces introduced to the system. The Greedy Pickup algorithm was an impactful heuristic to add to the collection process, since it would minimize the error of drift by re-identifying the piece of trash, allowing for it to not be marked exactly on the map. Compared to without the greedy pickup algorithm, the system was not effective since the original trash points marked in the map were never completely accurate and the was always some error in where the UGV estimated itself to be in the map.

%% file: sections/6_conclusion.tex
The presented autonomous multi-robot system is tailored for roadside litter monitoring and management. This paper presents an economical and scalable system that employs refined algorithms to compensate for imprecisions inherent in low-cost sensors. System integration of algorithms, including AMCL, DWA, Orbslam-2, and YOLO.v4, are harnessed to navigate, map, and detect trash respectively. Our Greedy Pickup algorithm, and anti-clustering filter, combined with a custom-designed collection mechanism, address trash identification errors, leading to an enhanced collection process.

The next steps for this system would to be refine the mechanical design of the UGV, such as adding odometry sensors onto the collection mechanism’s wheels and improving the wheels to be able to drive over more difficult terrains. There are also improvements to be made to the Greedy Pickup algorithm such as adding a Kalman filter to better approach trash pieces.

%% file: sections/7_acknowledgments.tex
\href{https://youtu.be/cIwk0Kh9k7E}{This} is a video of a full run-through our system did with two pieces of trash in the environment. \href{https://github.com/Capstone-W3/trash_parent_repo}{This} is the central repository in our organization that sets up the system. \href{https://github.com/Capstone-W3/caveman_mode}{This} is the Greedy Pickup repository which holds all the functions laid out in this paper. We would like to thank the contributions of Divya Venkatraman, Jared Raines and Catherine Ellingham for their work on gathering data, editing the paper, and system networking. 

%% file: sections/8_references.tex
\printbibliography